\documentclass[sigconf]{acmart}

\usepackage[normalem]{ulem} 

\usepackage{listings}
\usepackage{booktabs}
\usepackage{multirow}
\usepackage{algorithm}
\usepackage{algpseudocode}
\usepackage{amsmath}

\usepackage{ulem}
\usepackage{amsthm}
\usepackage{makecell}

\usepackage{subcaption}

\usepackage[most]{tcolorbox}
\usepackage{xcolor}
\newtcolorbox{promptbox}{
  colback=blue!5!white, 
  colframe=blue!75!black, 
  rounded corners,
  boxrule=0.5mm,
  boxsep=0.5mm, 
  left=2mm,
  right=2mm,
  top=0mm,
  bottom=0mm
}
\usepackage{balance}

\newcounter{noteZZctr} 

\newcounter{notePBctr} 

\AtBeginDocument{%
  \providecommand\BibTeX{{%
    \normalfont B\kern-0.5em{\scshape i\kern-0.25em b}\kern-0.8em\TeX}}}

\copyrightyear{2024}
\acmYear{2024}
\setcopyright{acmlicensed}
\acmConference[CIKM '24] {Proceedings of the 33rd ACM International Conference on Information and Knowledge Management}{October 21--25, 2024}{Boise, ID, USA.}
\acmBooktitle{Proceedings of the 33rd ACM International Conference on Information and Knowledge Management (CIKM '24), October 21--25, 2024, Boise, ID, USA}
\acmISBN{979-8-4007-0436-9/24/10}
\acmDOI{10.1145/3627673.3679830}

\begin{document}

\title{Distilling Large Language Models for Text-Attributed Graph Learning}

\author{Bo Pan}
\orcid{0009-0005-7501-7581}
\affiliation{
\institution{Emory University}
\department{Department of Computer Science}
\streetaddress{400 Dowman Drive}
\city{Atlanta}
\state{GA}
\postcode{30322}
\country{USA}
}
\email{bo.pan@emory.edu}

\author{Zheng Zhang}
\orcid{11111}
\affiliation{
\institution{Emory University}
\department{Department of Computer Science}
\streetaddress{400 Dowman Drive}
\city{Atlanta}
\state{GA}
\postcode{30322}
\country{USA}
}
\email{zheng.zhang@emory.edu}

\author{Yifei Zhang}
\orcid{11111}
\affiliation{
\institution{Emory University}
\department{Department of Computer Science}
\streetaddress{400 Dowman Drive}
\city{Atlanta}
\state{GA}
\postcode{30322}
\country{USA}
}
\email{yifei.zhang2@emory.edu}

\author{Yuntong Hu}
\orcid{0000-0003-3802-9039}
\affiliation{
\institution{Emory University}
\department{Department of Computer Science}
\streetaddress{400 Dowman Drive}
\city{Atlanta}
\state{GA}
\postcode{30322}
\country{USA}
}
\email{yuntong.hu@emory.edu}

\author{Liang Zhao}
\orcid{0000-0002-2648-9989}
\affiliation{
\institution{Emory University}
\department{Department of Computer Science}
\streetaddress{400 Dowman Drive}
\city{Atlanta}
\state{GA}
\postcode{30322}
\country{USA}
}
\email{liang.zhao@emory.edu}
\renewcommand{\shortauthors}{Bo Pan, Zheng Zhang, Yifei Zhang, Yuntong Hu, \&Liang Zhao}

\begin{abstract}
Text-Attributed Graphs (TAGs) are graphs of connected textual documents. Graph models can efficiently learn TAGs, but their training heavily relies on human-annotated labels, which are scarce or even unavailable in many applications. Large language models (LLMs) have recently demonstrated remarkable capabilities in few-shot and zero-shot TAG learning, but they suffer from scalability, cost, and privacy issues. Therefore, in this work, we focus on synergizing LLMs and graph models with their complementary strengths by distilling the power of LLMs into a local graph model on TAG learning. To address the inherent gaps between LLMs (generative models for texts) and graph models (discriminative models for graphs), we propose first to let LLMs teach an interpreter with rich rationale and then let a student model mimic the interpreter's reasoning without LLMs' rationale. We convert LLM's textual rationales to multi-level graph rationales to train the interpreter model and align the student model with the interpreter model based on the features of TAGs. Extensive experiments validate the efficacy of our proposed framework.
\end{abstract}





\maketitle

\section{Introduction}
Text-attributed graphs (TAGs) are a type of graph where each node is associated with a text entity, such as a document, and edges reflect the relationships between these nodes. TAGs harness the power of containing both semantic content and structural relations and thus have been predominantly utilized across various domains, including citation networks, e-commerce networks, social media, recommendation systems, and web page analytics, etc. \cite{zhao2022learning, zhu2021textgnn} The exploration of TAGs has been attracting significant interest due to its potential to transcend the conventional analysis of independent and identically distributed (i.i.d.) text features and focus on the relationship of text features. Current research on TAG learning typically adopts the pipeline of first extracting the text representations with a language model (LM), then feeding extracted text representation into a Graph Neural Network (GNN) to extract node embeddings with structural information \cite{chien2021node, zhao2022learning, duan2023simteg, zhu2021textgnn}. Such a pipeline allows the simultaneous capture of semantic and structural insights, yielding effective learning on TAGs.
However, the training of GNN typically heavily relies on annotated labels, which are tedious to prepare and may not be available in numerous tasks in TAGs \cite{wang2021zero, yue2022dual}. 

The recent emergence of Large Language Models (LLMs) brought light to solving the difficulty of data scarcity. For TAG learning, LLMs have proven zero-shot capabilities \cite{chen2023label, chen2023exploring}, and even become new state-of-the-art on some datasets \cite{huang2023can}. 
Despite the promise, the deployment, fine-tuning, and maintenance of LLMs require excessive resources, which may not be afforded by most of the institutions whose devices are not powerful enough, largely limiting their applicability. The cost of using public LLM APIs, such as ChatGPT, can be huge, especially for the TAG problem, which requires subgraphs of documents as the inputs \cite{wan2023efficient}. The practical applicability is further deteriorated by the privacy concerns of transferring sensitive data (e.g., in the network of health, social media, and finance, etc.) to public LLMs APIs \cite{yao2023survey}. These issues—scalability, cost, and privacy—underscore the necessity for a localized graph model that retains the advanced capabilities of LLMs without their associated drawbacks.

Given the need to have a localized model that also enjoys LLMs' power on TAG learning, a straightforward idea is knowledge distillation. 
For \textbf{language} models, previous research succeeded in distilling LLMs into smaller models and achieved comparable performance on specific tasks. Recent research \cite{ho2022large, hsieh2023distilling, li2023symbolic} found that jointly distilling the answers and rationales can be more effective than only using answers for distillation. However, the distillation of LLMs becomes a non-trivial task when the student models become \textbf{graph} models, especially when consider leveraging the rationales, due to the following challenges: 1) \textit{How to let language models teach graph models?} 
Language models are ``eloquent’’ teachers, which are typically generative models that output expressive information; while graph models, such as graph neural networks, have very succinct inputs and outputs (e.g., class labels), which may limit the knowledge absorption if using ordinary knowledge distillation by aligning the outputs. It is challenging yet important to make graph models sufficiently absorb expressive knowledge during training, while still can seamlessly adapt to succinct input and output when predicting. 2) \textit{How to transfer text rationales to graph rationales?} 
The rationales provided by LLMs are textual data that use natural language to explain the reasoning process, while graph rationale focuses on the salient areas in graphs most important to prediction. Seamlessly transferring across these heterogeneous rationales is seriously under-explored and challenging. 
3) 
\textit{How to synergize text and graph information during knowledge distillation?} Text-attributed graphs encompass the synergy of textual and graph topology information which are highly heterogeneous. How to ensure that both of these two types of knowledge as well as their interplay can be well preserved after knowledge distillation is a difficult and open problem.

To tackle these challenges, we propose a novel framework to distill LLMs into graph models. Our approach leverages the expressive outputs of LLMs during training through a two-stage distillation process. First, instead of having LLMs directly teach the student graph model, we introduce an \emph{interpreter} model that has a similar structure to the student model but can take the expressive outputs of LLMs as input. The student model is then aligned with the interpreter model that infers using raw data without rationales, enabling it to infer when LLMs are not available at test time.
To ensure the interpreter model is well-trained, we convert the LLM's textual rationales into enhanced features at the text-level, structure-level, and message-level, which inform its predictions. For effective alignment between the student and interpreter models, we propose a new TAG model alignment method that considers the feature discrepancies between text and graph embeddings. This approach ensures that the student model inherits the knowledge from the interpreter and can perform inference without relying on the expressive rationales from LLMs.

Our contributions can be summarized as follows:
\begin{itemize}
    \item We propose a new framework for distilling LLMs' knowledge to graph models for TAG learning. Our proposed framework achieves such knowledge distillation by letting LLMs output rationales to train an interpreter model, which is then aligned with a student model with no dependency on LLMs.
    \item We propose to convert textual rationales to text-level, structure-level, and message-level graph rationales as enhanced features for the interpreter model, and LLM-generated pseudo-labels and pseudo-soft labels as supervision to train the interpreter model.
    \item We propose a semantics and structure-aware TAG model aligning method. The proposed alignment method preserves the text and graph information in aligning TAG models, thus allowing the student model to better align with the interpreter model.
    \item We conducted comprehensive experiments to validate the performance of our proposed framework. The proposed method consistently beat the baseline methods by an average improvement of 6.2\% across four datasets. 
\end{itemize}


\section{Related Work}\label{sec:related}
\subsection{LLMs for Text-Attributed Graphs Learning}
Current research on applying LLMs to help TAG learning contains two categories: LLM-as-predictor and LLM-as-enhancer. The first stream of methods starts with the research of directly utilizing LLMs for zero-shot prediction \cite{chen2023label, chen2023exploring}. Research shows LLMs even become new state-of-the-art on some datasets \cite{huang2023can, hu2023beyond}. A new focus of recent research is tuning LLMs for better prediction on graphs \cite{ye2024language, chen2024llaga, tang2023graphgpt, zhang2024taga}. The second stream of work, LLM-as-enhancer, focuses on using LLMs to enhance the features of TAG. For example, LLM has been proven the power of node feature enhancement \cite{he2023explanations}, edge editing \cite{sun2023large}, etc. Such power can greatly benefit the learning on TAGs by providing augmented data and features for training the models \cite{he2023explanations, sun2023large}. 
Although LLMs have superior power of text feature understanding or local link modification, they still lack the ability of capturing higher-hop neighbor information due to the constraints on input text length, and such ability is the advantage of GNNs. Therefore, using LLMs to enhance the features and combining them with GNNs has become a new state-of-the-art paradigm in TAG learning \cite{he2023explanations, sun2023large, chen2023label}. However, both these two streams of work requires calling LLMs during the test time, which has the issues of cost and privacy. How to benefit from LLMs in the training but predict without LLMs during test time is still an open problem.
\subsection{Knowledge Distillation of LLMs}
Previous research in the NLP domain succeeded in distilling larger-scale pre-trained language models into smaller models and achieving comparable performance~\cite{sanh2019distilbert, jiao2019tinybert, dasgupta2023cost}. In the LLM's era, recent work found distilling the rationales as well as answers generated by LLMs in a multi-task learning setting can significantly improve the performance of the student model \cite{magister2022teaching, ho2022large, hsieh2023distilling, li2022explanations, shridhar2023distilling, zhang2024elad, bai2024beyond}. Specifically, when distilling LLMs to smaller language models, the rationales like explanations \cite{hsieh2023distilling} or chain-of-thought \cite{shridhar2023distilling} are also used as supervision for the student model to better learn the reasoning capabilities.
When the target models are graph models, one work attempted to use LLM as an annotator for node classification \cite{chen2023label}, but it only considered distilling the predictions and failed to leverage the rationales of LLMs. To the best of our knowledge, no existing work has attempted to leverage the rationales to help the knowledge distillation into graph models.
\subsection{Privileged Information} 
Privileged Information \cite{vapnik2015learning} introduces a teacher who provides additional information to a student model during the learning process. The underlying idea is that the teacher’s extra explanations can help the student develop a more effective model. 
\textit{Generalized distillation} \cite{lopez2015unifying} is a framework that unifies privileged information and knowledge distillation \cite{hinton2015distilling, zhang2023visual}. It gives a more general form of knowledge distillation that allows the teacher model to have privileged input information than the student model. The generalized distillation framework first learns a teacher model with the data where the input is privileged information and generates a set of soft labels for each data. Then the student model is learned with the original features by aligning the soft labels with the teacher model. 

\begin{figure*}[ht]
    \centering
    \includegraphics[width=1\linewidth, page=1]{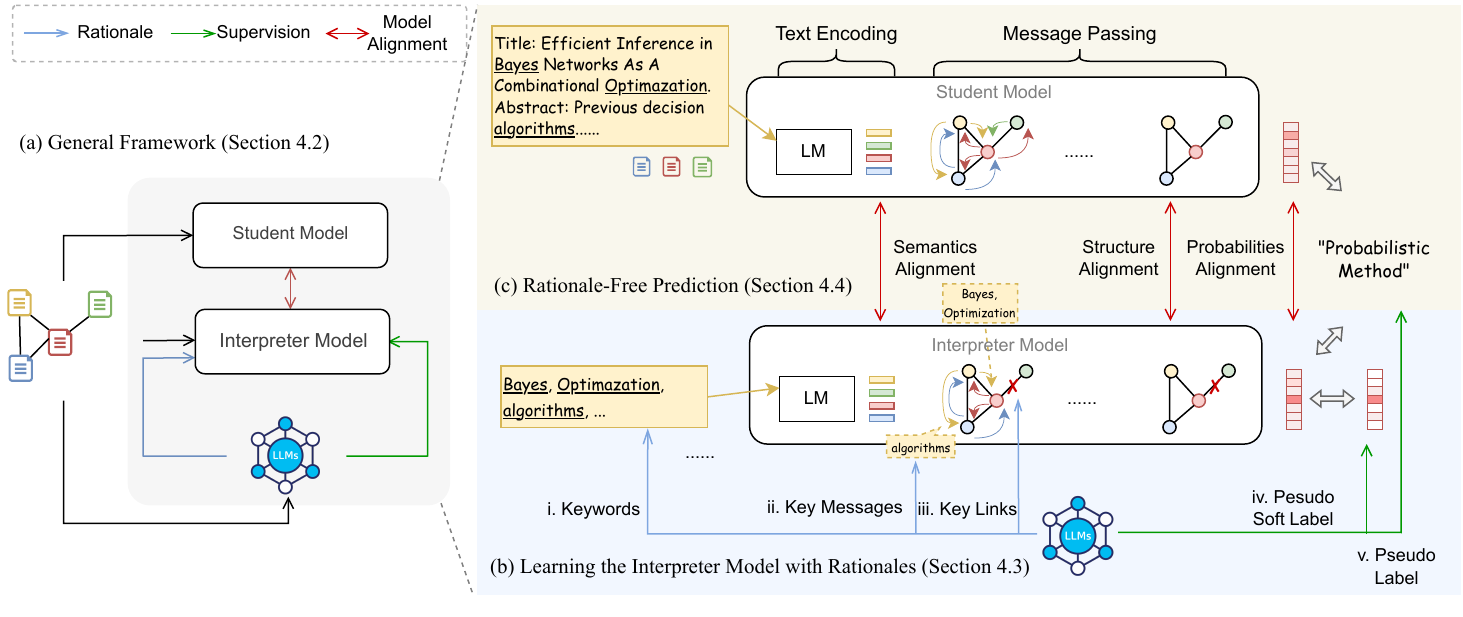}
    \caption{Illustration of our proposed LLM to graph model knowledge distillation framework. (a) The general distillation framework. We propose to distill the knowledge of an LLM by leveraging the LLM-generated rationales and supervision to train an interpreter model, then align the student model on raw features with the interpreter model. (b) The training of the interpreter model with rationales and pseudo-supervision. The interpreter model takes rationales as input, including keywords, key edges, and key messages. LLM-generated pseudo supervision, including pseudo-label and soft labels, is used to train the interpreter model. (c) The proposed model alignment framework. The student model which takes original features as input is aligned with the interpreter model on text and graph levels based on the discrepancy between raw inputs and rationale-enhanced inputs.}
    \label{fig:method}
\end{figure*}

\section{Preliminaries}\label{sec:preliminary}
\textbf{Learning on Text-Attributed Graphs.} 
TAG learning aims to learn representations of graphs, each of whose nodes is associated with a text feature. Most existing work on TAG learning follows the paradigm of first using a pre-trained language model (LM) to encode the text features into embeddings, then using a graph neural network (GNN) to aggregate the neighbor information to get the node embedding. This approach leverages the strengths of both language models in understanding textual information and GNNs in capturing the relational structure of the graph. In this work, we also follow this paradigm. 

Formally, a TAG can be represented as $\mathcal{G}=\left(\mathcal{V}, A,\mathcal{X}\right)$, where $\mathcal{V}=\{v_0, v_1, ..., v_{N-1}\}$ is a set of $N$ nodes, $A \in \{0,1\}^{N \times N}$ is the adjacency matrix, and $\mathcal{X}=\{x_0, x_1, ..., x_{N-1}\}$ is the set of text features where $x_n$ is the text feature associated with node $v_n \in \mathcal{V}$. 
For a node $v_n$, its initial text embedding is obtained by 
\begin{equation}
    h_{n,0} = \text{LM}(x_n)
\end{equation}
where $\text{LM}$ is the text encoder, usually implemented with a BERT-like language model, and $x_n$ is the text associated with node $v_n$.
After obtaining the text embeddings as initial node embeddings, a GNN adopts message passing to learn the structure-aware embeddings for each node. A general message-passing operation can be expressed as:
\begin{equation}\label{eq:attention_based_message}
\begin{split}
& h_{n,k} = \text{UPD}\big(h_{n,k-1},\\ 
& \text{AGG}\left(\{\text{MSG}(h_{n,k-1},h_{m,k-1})\}_{ v_m \in \mathcal{N}(v_n)}\right)\big) 
\end{split}
\end{equation}
where $h_{n,k}$ represents the embedding of node $v_n$ in the $k-$th layer of the graph neural network, $\mathcal{N}(v_n)$ is the set of neighbor nodes of $v_n$. The $\text{AGG}$ function is used for aggregating the neighbor node embeddings and $\text{UPD}$ is the function to update the embedding of $v_n$ based on the aggregated neighbor node embedding. 
Finally, the node classification is made with the final node representation by 
$\hat{t}_n=MLP(h_{n,l})$
where $l$ is the number of message passing layers and $\hat{t}_n$ is the predicted soft labels (logits).

\section{Methodologies} \label{sec:method}
\subsection{Problem Formulation}\label{sec:problem}

In this work, we delve into the task of distilling an LLM to a local model for TAGs. We focus on the foundational task of node classification. Given a text-attributed graph $\mathcal{G}=\left(\mathcal{V}, A,\mathcal{X}\right)$ and a large language model $LLM$, the goal is to train a student model, which consists of a language model $LM^\mathcal{S}$ and a graph neural network model $GNN^\mathcal{S}$ with the capability to infer using the features $\mathcal{V}, A,\mathcal{X}$. During the learning process, no ground truth labels are provided, and only a subset of nodes $\mathcal{V}_{train}$ and their textual features $\mathcal{X}_{train}$ are allowed to be exposed to LLMs.

This problem presents several unique challenges, especially when considering leveraging LLM's rationales in the distillation, due to the following reasons: 
\begin{itemize}
    \item \textbf{Difficulty in leveraging language models to teach graph models.} The teacher model (LLM) and the student model (graph models) differ significantly in their model architectures and output forms. LLMs generate succinct text information, while the output of the student graph models are merely class labels. Traditional knowledge distillation by aligning the logits and parameters cannot make the student model fully absorb the knowledge from the teacher model. 
    \item \textbf{Difficulty in transferring text rationales to graph rationales.} LLMs can provide textual rationales for TAG learning, which the graph models cannot naturally understand. Transferring the text rationales to graph rationales is a challenging yet under-explored problem.
    \item \textbf{Difficulty in synergizing text and graph information in model distillation.} In the distillation process, the student model should be well aligned with the teacher model for a better understanding of both the text and structure information. It is vital but challenging to preserve text and graph knowledge and their interplay in the distillation.
\end{itemize}

\begin{figure}[t]\centering
\vspace{-6mm}
\includegraphics[width=1\linewidth]{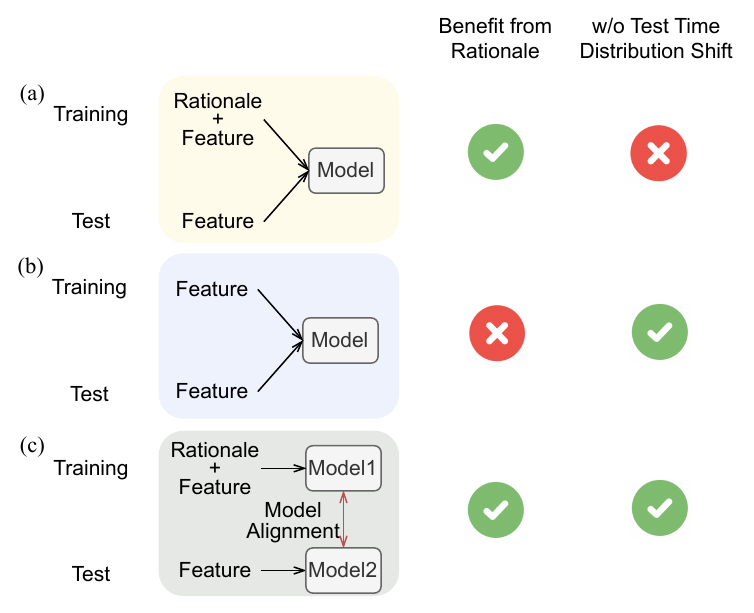}
\vspace{-3mm}
    \caption{Different approaches to incorporate rationales to train graph models.}
    \label{fig:framework}
\vspace{-6mm}
\end{figure}

\subsection{A General Framework for Distilling LLM to Graph Models} \label{sec:framework}
To tackle the difficulty in leveraging language models to teach graph models, we propose a framework that bridges the gap between these two types of models by enabling the student graph model to comprehend and apply the knowledge derived from LLMs.

LLMs can provide rich knowledge on TAG learning, i.e., rationales. The rationales can be depicted as the important nodes, edges, and text features for each prediction. Graph models, which are discriminative models, can benefit from the rationales by taking them as input. Taking the rationales as input helps the model make predictions easier since the information in the rationale is exactly leading to the prediction. To make use of LLM's knowledge, a naive idea is to use LLM-provided rich knowledge as additional features to train the graph model (Fig.~\ref{fig:framework}~(a)). However, although the rationales can benefit the training process, using them as inputs to train the model leads to even worse test performance due to the distribution shift of features at the test time \cite{wu2022knowledge}. An alternative solution is to only leverage succinct knowledge (i.e. LLM's predictions) as supervision to train the model so that the training and test features can be in the same distribution (Fig.~\ref{fig:framework}~(b)). However, this method falls short of leveraging the full rationale power of LLMs, which encompasses rich information crucial for the student model's understanding of the predictions. 

To address this problem and ensure the model can generalize to original features without rationales at test time, we propose a solution which is first to use LLMs' rationale to train an intermediate model, then use the intermediate model to train the student model with the same architecture but takes raw features without rationales (Fig.~\ref{fig:framework}~(c)).

Leveraging this idea in knowledge distillation, we propose a novel framework for the LLM-to-graph model knowledge distillation problem, illustrated in Fig.~\ref{fig:method}~(a). Our proposed framework involves an intermediate \textit{Interpreter} model, which bridges the LLM and the student model. 
Specifically, to extract the knowledge of LLMs on TAG learnings, we let LLMs provide rationales (the blue arrow) to enhance the features and provide supervision (the green arrow) to train the model. Since the interpreter model and the student model share a similar model structure, the student model can be aligned with the interpreter model by aligning their latent embeddings (the red arrow). 

Given this general framework, several specific questions arise that need to be addressed to enhance its practical application. These include: (1) how to incorporate LLM-generated textual rationales into the interpreter graph model, and (2) how to effectively align the student model with the interpreter model. The remainder of this section addresses these questions, with the first question discussed in Section~\ref{sec:teacher} and the second in Section~\ref{sec:distill}.
\subsection{Interpreter Model: Zero-Shot TAG Learning with Rationales} \label{sec:teacher}
To tackle the difficulty in transferring text rationales to graph rationales, in this section we propose a new method to convert TAG rationales into multiple-level enhanced features. In this part, the goal is to train a strong interpreter model, consisting of $LM^\mathcal{T}$ and $GNN^\mathcal{T}$. The rationales behind a TAG model decision can be depicted as:

\textit{A prediction is primarily made based on which part(s) of text on which neighbor node(s)}.

Since the graph models cannot directly take the textual rationales, to bridge the style of textual and graph rationales, we convert the textual rationale into three forms of feature enhancements that help with the prediction of the answer: i) keywords in texts, ii) key links around the central node,
and iii) key semantic messages from the neighbors. To operationalize the training, we leverage LLMs to generate supervision to train the interpreter model, including iv) pseudo-soft labels and v) pseudo-labels. The training of the interpreter model with rationales is illustrated in Fig.~\ref{fig:method}~(b).

We prompt such prediction and reasoning in a three-step manner: 1) We input the text attributes of each node, along with those of its neighbors, into LLMs to generate pseudo-labels and pseudo-soft labels.
2) We then supply the node's text and its pseudo-label to LLMs to identify critical keywords within the node's attributes for the classification.
3) We feed the text of the node and its neighbors along with its pseudo-label into LLMs to identify essential links and messages for the classification.
Each of these steps will be introduced in detail in the following. 

\textbf{Pseudo-Label and Soft Label Creation}. Since soft labels can have more information compared to hard categories in knowledge distillation \cite{lopez2015unifying}, we first leverage the zero-shot abilities of LLMs to generate the pseudo-labels and pseudo-soft labels to assist in the training of the interpreter model. The generated labels are also used as the targeted answers for generating rationales. The process of label and soft-label creation can be written as
\begin{equation}
    l_n, y_n=LLM(x_n, \{x_i\}_{\{i|v_i\in\mathcal{N}(v_n)\}}; prompt_{L})
\end{equation}
where $y_n$ and $l_n$ are the LLM-predicted pseudo-label and soft label for node $v_n$, respectively. $LLM(x_n, \{x_i\}_{\{i|v_i\in\mathcal{N}(v_n)\}}; prompt_{L})$ means calling LLM with the prompt $prompt_{L}$ and variables $x_n$ and $\{x_i\}_{\{i|v_i\in\mathcal{N}(v_n)\}}$. A simplified example of $prompt_{L}$ is given below for easier understanding. Note that the format conversion from textual outputs to numerical values is omitted in all the mathematical expressions.\footnote{All the formatting instructions in prompts are omitted.}

\begin{tcolorbox}[colback=blue!5!white,colframe=blue!50!black,title={A simplified example of  $prompt_{L}$}]
\small 
\textbf{Input:}We want to classify a paper into the following categories: [Neural Networks, Generic Algorithms, ...]. Please identify logits-like probabilities for each class and give your final classification. The paper is ... Its neighbors are ...\\
\textbf{Response:} \{Probabilities: [0.05, 0.85, ...], Category: 'Generic Algorithm'\}
\end{tcolorbox}

\textbf{Keyword Recognition}.
LLMs have been proven to have the capability to extract keywords in languages \cite{hajialigol2023xai}. In the interpreter model, we leverage the reasoning power of LLMs to extract the important keywords that are most helpful for the classification of the text to the predicted class, thus removing the words that can mislead the classification. We then concatenate the keywords and feed them to the local $LM$ to extract text embeddings. Formally, assuming $prompt_{KW}$ is the prompt for keyword recognition, for a node $v_n$ with a text attribute $x_n$, we feed it to an LLM with $prompt_{KW}$ to obtain the recognized keywords. The enhanced text embedding can be obtained as
\begin{equation}\label{eq:LM-inter}
h_{n,0}^\mathcal{T}=LM^\mathcal{T}\left([LLM(x_n, y_n; prompt_{KW})]\right)
\end{equation}
where $LLM(x_n, y_n; prompt_{KW})$ means utilizing LLMs to extract a list of important keywords of $x_n$ that helps to classify it to $y_n$ with the prompt $prompt_{KW}$. Here the output of the function $LLM$ with $prompt_{KW}$ is seen as the set of keywords. $[\cdot]$ denotes concatenating the keywords with spaces as the separator. A simplified example of $prompt_{KW}$ is given below.
\begin{tcolorbox}[colback=blue!5!white,colframe=blue!50!black,title={A simplified example of  $prompt_{KW}$}]
\small 
\textbf{Input:} We want to classify a paper into the following categories: [Neural Networks, Generic Algorithms, ...]. Please identify at most 5 words in the provided text that help most with the classification to 'Neural Networks'. The paper is ... Its neighbors are ...\\
\textbf{Response:} [Algorithm, Optimization, Bayesian]
\end{tcolorbox}

\textbf{Key Links and Messages Recognition}. For structural reasoning, we prompt LLM to identify the key links around a specific central node by identifying a subset of neighbors (key links), and also the keywords in each of the neighbor nodes (key messages), that are important for the central node's classification to the predicted class. The identified key links are used as edited local structures in message passing, and the key messages are used to replace the messages in the first graph convolutional layer. Formally, for a central node $v_n$, we provide $LLM$ with its text attribute $x_n$, its neighbors' texts $ \{x_i\}_{\{i|v_i\in\mathcal{N}(v_n)\}}$, the pseudo-label $y_n$ and the prompt for this task $prompt_{KL}$, and extract the important neighbor set and keyword set from LLM's response. For expression simplicity, we write them as the function $LLM$'s two outputs with $prompt_{KL}$. The key links and message recognition are written as
\begin{equation}
\begin{aligned}
&\mathcal{N}'(v_n),\{{w}_{n,i}\}_{\{i|v_i\in\mathcal{N'}(v_n)\}} \\ & = LLM\left(x_n,  \{x_i\}_{\{i|v_i\in\mathcal{N}(v_n)\}}, y_n; prompt_{KL}\right)    
\end{aligned}
\end{equation}
where $\mathcal{N}'(v_n)$ is the key neighbor nodes and ${w}_{n,i}$ is the keywords in the message from node $v_i$ to the central node $v_n$. With the enhanced structural features, the message passing is done by
\begin{equation}\label{eq:gnn-inter}
\begin{aligned}
h_{n,k}^{\mathcal{T}}& = \text{UPD}\big(h_{n,k-1}^{\mathcal{T}}, \\ 
&\text{AGG}(\{\text{MSG}(h_{n,k-1}^{\mathcal{T}},h_{m,k-1}^{\mathcal{T}}) | v_m \in \mathcal{N}'(v_n)\})\big)    
\end{aligned}
\end{equation}
The messages in the first graph convolutional layer are replaced with 
\begin{equation}
\text{MSG}(h_{n,0}^{\mathcal{T}},h_{i,0}^{\mathcal{T}})=LM^\mathcal{T}\left([w_{n,i}]\right)
\end{equation}
where $[\cdot]$ denotes concatenating the words in the key message from $v_i$ to $v_n$. A simplified example of $prompt_{KL}$ is given as below.
\begin{tcolorbox}[colback=blue!5!white,colframe=blue!50!black,title={A simplified example of  $prompt_{KL}$}]
\small 
\textbf{Input:} We want to classify a paper in a citation network to the following categories: [Neural Networks, Generic Algorithms, ...] Please identify a subset of important neighbors and at most 5 keywords of each important neighbor that help most to classify the central node into the category of 'Neural Networks'. The paper is ... Its neighbors are ...\\
\textbf{Response:} \{Node 0: ['Bayesian', 'Learning'], Node 248: ['Optimize', 'Convex']\}
\end{tcolorbox}

\textbf{Training Objective of the Interpreter Model.}
With the above LLM-provided rationales, the learning of TAG is achieved with Eq.~\ref{eq:LM-inter} ($LM^\mathcal{T}$) and Eq.~\ref{eq:gnn-inter} ($GNN^\mathcal{T}$).
After obtaining the last layer's node embedding $h_{n,l}^\mathcal{T}$ with Eq.~\ref{eq:gnn-inter}, the prediction of the interpreter model is made by $\hat{t}_n^\mathcal{T}=MLP^\mathcal{T}(h_{n,l}^\mathcal{T})$, where $\hat{t}_n^\mathcal{T}$ is the predicted soft label. The interpreter model is trained to predict the label, and soft labels in a multi-task learning manner. The overall loss function to train the interpreter model is given by:
\begin{equation}
\begin{aligned}
     l^{\mathcal{T}}=\hspace{-2mm}\sum_{\{n\}_{v_n \in \mathcal{V}_{train}}} \hspace{-2mm}\bigg(l_{label}^{\mathcal{T}}(\hat{t}_n^\mathcal{T}, y_n)+\lambda_1 l_{logits}^{\mathcal{T}}(\hat{t}_n^\mathcal{T}, l_n)\bigg)
\end{aligned}    
\end{equation}

where $l_{label}$ is implemented with the Cross-Entropy (CE) loss, $l_{logits}$ is implemented with the Mean Square Error (MSE) loss, and $\lambda_1$ is a hyper-parameter.
\subsection{Semantics and Structure-Aware TAG Model Alignment}\label{sec:distill}
To achieve better LLM-free prediction at test time, we propose a new model alignment method for TAGs, as illustrated in Fig.~\ref{fig:method}~(c). The proposed method jointly considers the textual embeddings (LM-extracted) and structural embeddings (GNN-extracted) between the interpreter and student models, thus allowing the student model to give more similar predictions to the interpreter model but does not need to take rationales as input.

\textbf{Semantics Alignment.} For semantic representation, we extract text embeddings from the interpreter model and student model with weights considering their occurrence frequency in the graph structures. We also focus more on those nodes whose LLM-extracted keywords are more different from their raw text when aligning them. Formally, the semantic alignment loss can be written as:
\begin{equation}
    l_{semantic}^{\mathcal{S}} = \sum_{\{n\}_{v_n \in \mathcal{V}_{train}}} \frac{\text{Deg}(v_n) }{\text{sim}_t(x_{n}, x^\mathcal{T}_{n})}\cdot d(h^{\mathcal{S}}_{n,0}, h^\mathcal{T}_{n,0})
\end{equation}
where $\text{Deg}(v_n)$ is the degree of $v_n$, $\text{sim}_t(x_{n}, x^\mathcal{T}_{n})$ is the semantic similarity between the raw text feature and the LLM-enhanced text feature of node $v_n$, which can be calculated with the cosine similarity of their embeddings extracted from a pre-trained language model like Bert, and $d(h^{\mathcal{S}}_{n,0}
, h^\mathcal{T}_{n,0})$ is the distance between their text embeddings that we aim to minimize. 


\textbf{Structure Alignment.}
In structural alignment, we also focus more on those nodes with more discrepancy between the original structure and the rationale-enhanced structure. Since in our proposed method, the structure enhancement is done by selecting important neighbors from all neighbors, the discrepancy can be calculated as the difference of neighbor node numbers. Formally, the structural alignment loss can be written as:
\begin{equation}
l_{structural}^{\mathcal{S}} = \hspace{-4mm}\sum_{\{n\}_{v_n \in \mathcal{V}_{train}}}\hspace{-5mm}\frac{\text{Deg}(v_n)}{\text{sim}_s(\mathcal{N}(v_n),\mathcal{N}'(v_n))} \cdot  d(h^{\mathcal{S}}_{n,l}, h^\mathcal{T}_{n,l})    
\end{equation}
where $\text{sim}_s(\mathcal{N}(v_n),\mathcal{N}'(v_n))$ measures the similarity between $v_n$'s original neighbor structure and its enhanced neighbor structure, calculated by
\begin{equation}
\text{sim}_s(\mathcal{N}(v_n),\mathcal{N}'(v_n))=\frac{1}{|\mathcal{N}_k(v_n)|-|\mathcal{N}'_k(v_n)|}
\end{equation}
and $d(h^{\mathcal{S}}_{n,l}, h^\mathcal{T}_{n,l})$ is a measure of the distance of node embeddings, which we aim to minimize. 

\begin{table*}
\centering
\caption{Main Results of Zero-Shot Test-Time LLM-Free Node Classification. }
\label{tab:main}
\begin{tabular}{@{}ccccccc@{}}
\toprule
LLM's Role & Method  & Backbone & Cora & PubMed & ogbn-products & arxiv-2023 \\
\midrule
\multirow{14}{*}{LLM as Annotator} & \multirow{3}{*}{LM} & Bert & 0.7400±0.0175 & 0.9058±0.0046 & 0.7020±0.0033 & 0.6840±0.0122 \\
& & DistilBert & 0.7355±0.0163 & 0.9028±0.0053 & 0.7080±0.0040 & 0.6821±0.0100  \\
& & Deberta & 0.7385±0.0127 & 0.9020±0.0057 & 0.7074±0.0056 & 0.6789±0.0185\\
\cmidrule{2-7}
& \multirow{3}{*}{GNN (Shallow)} & GCN & 0.7126±0.0213 & 0.8322±0.0076 & 0.5593±0.0031 & 0.6355±0.0032 \\
& & GAT & 0.7186±0.0346 & 0.8122±0.0282 & 0.5583±0.0010 & 0.6351±0.0060 \\
& & SAGE &0.7149±0.0223 & 0.8287±0.0107 & 0.5475±0.0023 & 0.6460±0.0006 \\
\cmidrule{2-7}
& \multirow{3}{*}{GNN (PLM)} & GCN & 0.6720±0.0333 & 0.8136±0.0099 & 0.6944±0.0111 & 0.6647±0.0068 \\
& & GAT &0.6628±0.0434 & 0.7996±0.0229 & 0.7049±0.0043 & 0.6675±0.0059 \\
& & SAGE &  0.6619±0.0191 & 0.7968±0.0118 & 0.6879±0.0067 & 0.6748±0.0067 \\
\cmidrule{2-7}
& \multirow{3}{*}{GIANT} & GCN & 0.7205±0.0045 & 0.8122±0.0048 & 0.6977±0.0042 & 0.6679±0.0067 \\
& & GAT & 0.7233±0.0024 & 0.8077±0.0079 & 0.7189±0.0030 & 0.6822±0.0073 \\
& & SAGE & 0.7145±0.0033 & 0.8202±0.0046 & 0.6869±0.0119  & 0.6683±0.0037 \\
\cmidrule{2-7}
& \multirow{1}{*}{GLEM} & RevGAT & 0.7312±0.0017	& 0.7863±0.0122	& 0.7126±0.0059	& 0.6823±0.0030 \\
\cmidrule{2-7}
& \multirow{1}{*}{SIMTEG} & SAGE & 0.7327±0.0020	& 0.8327±0.0113 &	0.7037±0.0053	&0.7233±0.0034 \\
\midrule
\multirow{3}{*}{\parbox[c]{3cm}{\centering \textbf{LLM as Teacher} }} & \multirow{3}{*}{\textbf{(Proposed)}}& GCN & 0.8237±0.0187 & 0.9215±0.0096 & 0.7333±0.0025 & 0.7801±0.0424 \\
& & GAT & ${0.8237}$±${0.0137}$ & 0.9189±0.0019 & ${0.7346}$±${0.0030}$ & 0.7838±0.0424 \\
& & SAGE & 0.8210±0.0296 & ${0.9217}$±${0.0105}$ & 0.7283±0.0015 & ${0.7918}$±${0.0456}$\\
\bottomrule
\end{tabular}

\end{table*}

\textbf{Training Objective of Model Alignment.} In addition to the above two distillation loss functions, we also align the logits between the interpreter and the student model. The overall objective in model alignment can be written as:
\begin{equation}
l^{\mathcal{S}} = \underbrace{l_{label}^{\mathcal{S}}+\lambda_2 l_{logits}^{\mathcal{S}}}_{\text{standard distillation}} + \underbrace{ \lambda_3 l_{semantic}^{\mathcal{S}} + \lambda_4 l_{structural}^{\mathcal{S}}}_{\text{TAG model alignment}}
\end{equation}
where $l_{label}^{\mathcal{S}}$ is the prediction loss of the student model, which is usually calculated with cross-entropy loss. $l_{logits}^{\mathcal{S}}$ is the logits alignment loss between the interpreter and student model, which is usually implemented with the mean square error (MSE) loss. $\lambda_2, \lambda_3, \lambda_4$ are the hyper-parameters.

\section{Experiments}\label{sec:experiments}
\subsection{Experimental Setting} \label{sec:setting}
\textbf{Datasets.}
We evaluate our proposed framework on four text-attributed graph datasets: Cora \cite{mccallum2000automating}, Pubmed \cite{sen2008collective}, ogbn-products \cite{hu2020open} and arxiv-2023 \cite{he2023explanations}. Each dataset is described in detail as follows: \textbf{Cora} is a paper citation network dataset consisting of 2,708 scientific publications from the computer science domain. The shallow node attributes are represented by a 1,433-dimensional binary vector indicating the presence of specific words in the document. \textbf{Pubmed} is a paper citation dataset consisting of 19,717 scientific journals collected from the PubMed database, which focuses on the biomedical domain. Each publication is classified into one of three categories related to diabetes.  The node attributes are represented by a Term Frequency-Inverse Document Frequency (TF-IDF) weighted word vector with 500 unique words. \textbf{Ogbn-products} \cite{hu2020open} is an Amazon co-purchase network where each node represents a product. The corresponding input raw text consists of titles and descriptions of products. The node labels are categories of products, and there are 47 distinct categories. We use the same subset of ogbn-products as in \cite{he2023explanations}. Shallow embeddings are extracted with BoW. \textbf{Arxiv-2023} \cite{he2023explanations} is a dataset comprising papers published in 2023 or later. This dataset is specifically designed to address the concern of potential label leakage, as it includes papers that are beyond the knowledge cutoff for models like GPT-3.5. This dataset helps in evaluating the model's performance on more recent and unseen data. The shallow embeddings are extracted with word2vec.
The statistics of each dataset are given in Table~\ref{tab:datasets}. Following previous work \cite{he2023explanations}, for all datasets, we randomly split the training/validation/test set with the ratio of 60/20/20.

\begin{table}[H]
\centering
\caption{Statistics of the TAG datasets.}
\label{tab:datasets}
\begin{tabular}{lccccc}
\toprule
Dataset & \#Nodes & \#Edges & \#Classes \\
\midrule 
Cora & 2,708 & 5,429 & 7  \\
Pubmed & 19,717 & 44,338 & 3\\
ogbn-products & 54,025 & 74,420 & 47 \\
arxiv-2023 & 46,198 & 78,548 & 40 \\
\bottomrule
\end{tabular}
\end{table}

\textbf{Comparison Methods.} Since our problem setting is LLM-free prediction, we only compare to methods that is independent with LLMs at test time. For a fair comparison, we compare our proposed method with distilling the prediction without rationales (\textbf{LLM as Annotator}) to train the local models like LMs and GNNs. Specifically, our comparison methods include: 
\begin{enumerate}
    \item \textbf{LMs} (Language Models): Pre-trained language models fine-tuned by LLM-provided pseudo-labels are adopted as the first type of comparison method. Pre-trained Bert \cite{devlin2018bert}, DistilBert \cite{sanh2019distilbert} and DeBERTa \cite{he2020deberta} are tested as model backbones.
    \item \textbf{GNNs}: For GNNs, we tested GCN \cite{kipf2016semi}, GAT \cite{velivckovic2017graph} and GraphSAGE \cite{hamilton2017inductive} as backbones. We compare our method with shallow and PLM initial node features: 
    \begin{enumerate}
        \item \textbf{Shallow}: Text embeddings provided by the PyG library \cite{Fey/Lenssen/2019}, extracted with shallow methods including BoW, TD-IDF and word2vec, as introduced in each dataset's description.
        \item \textbf{PLM} : Text embeddings extracted from Bert-like Pre-trained Language Models are used as node features. DistilBert \cite{sanh2019distilbert} is used as the language encoder.
    \end{enumerate}
    \item \textbf{GIANT}: GIANT  \cite{chien2021node} is a self-supervised node feature extraction framework that aims to extract more efficient node features for text-attributed graphs. For fair benchmarking, LLM-provided pseudo-labels are used to fine-tune the text encoder of GIANT.
    \item \textbf{GLEM}: GLEM \cite{zhao2022learning} is an efficient TAG learning framework by fusing graph structure and language learning with a variational Expectation Maximization (EM) framework. LLM-provided pseudo-labels are used as labels to train GLEM. Following the original work, RevGAT is used as the GNN backbone.
    \item \textbf{SIMTEG}: SIMTEG \cite{duan2023simteg} is a TAG learning framework via supervised fine-tuning of the LM encoder of node features. Following the original work, GraphSAGE is used as the GNN backbone.
\end{enumerate}
For all methods, GPT-3.5 Turbo (1106) is used as annotators/teachers.

\textbf{Training and Implementation Details.} We follow the standard inductive learning setting \cite{zhang2021graph}, all links between the training nodes and test nodes in the graphs are removed in the training stage to ensure all test nodes are unseen during training. The implementation of the proposed method utilized the PyG and Huggingface libraries, which are licensed under the MIT License. All experiments were conducted on a single NVIDIA RTX 3090 GPU with 24GB VRAM. For pre-training language models, we use a max token number of 512 for full-text features and 48 for keyword features for all three pre-trained language models including bert-base-uncased, distilbert-base-uncased, and microsoft/deberta-base. For the GNN models including GCN, GAT and SAGE, we test on 2-layer models with a hidden dimension of 256. In the training process, we first train the interpreter model alone until convergence, then fix the interpreter model to train the student model.
The interpreter model is trained first, then we freeze it and train the student model. We adopt a learning rate of 1e-5 to fine-tune all the pre-trained language models and train for 10 epochs. We train all GNN models for 200 epochs with a learning rate of 0.01.

\begin{figure*}[!h]
    \centering
    \begin{subfigure}{0.24\textwidth} 
        \includegraphics[width=\linewidth]{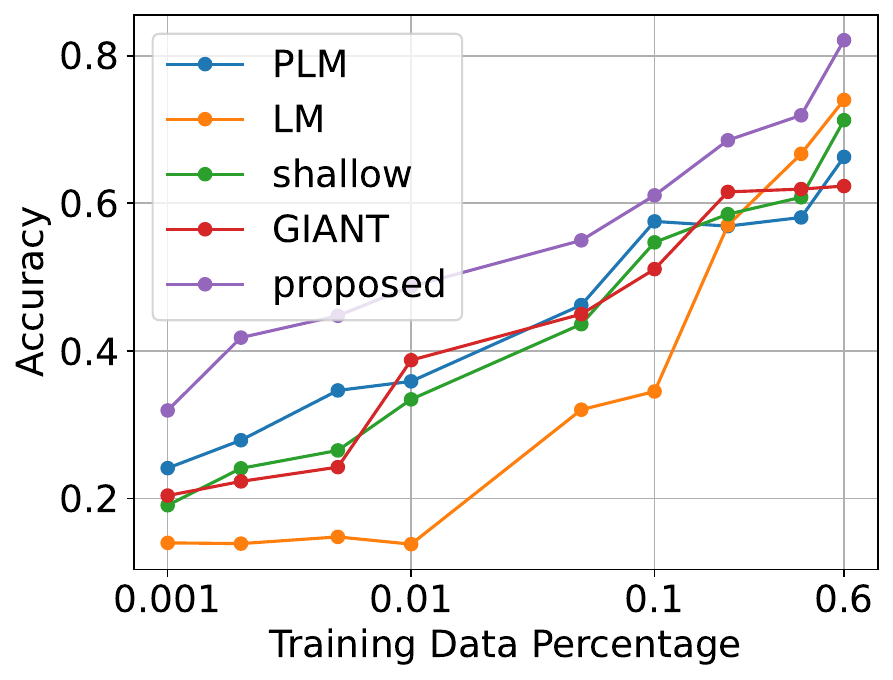}
        \caption{Cora}
    \end{subfigure}
    \begin{subfigure}{0.24\textwidth}
        \includegraphics[width=\linewidth]{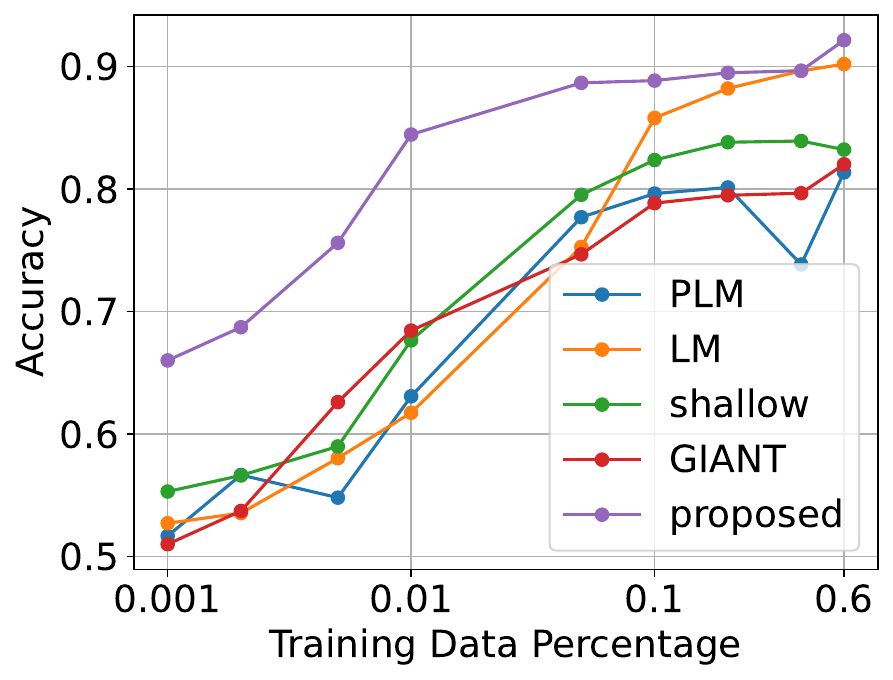}
        \caption{Pubmed}
    \end{subfigure}
    \begin{subfigure}{0.24\textwidth}
        \includegraphics[width=\linewidth]{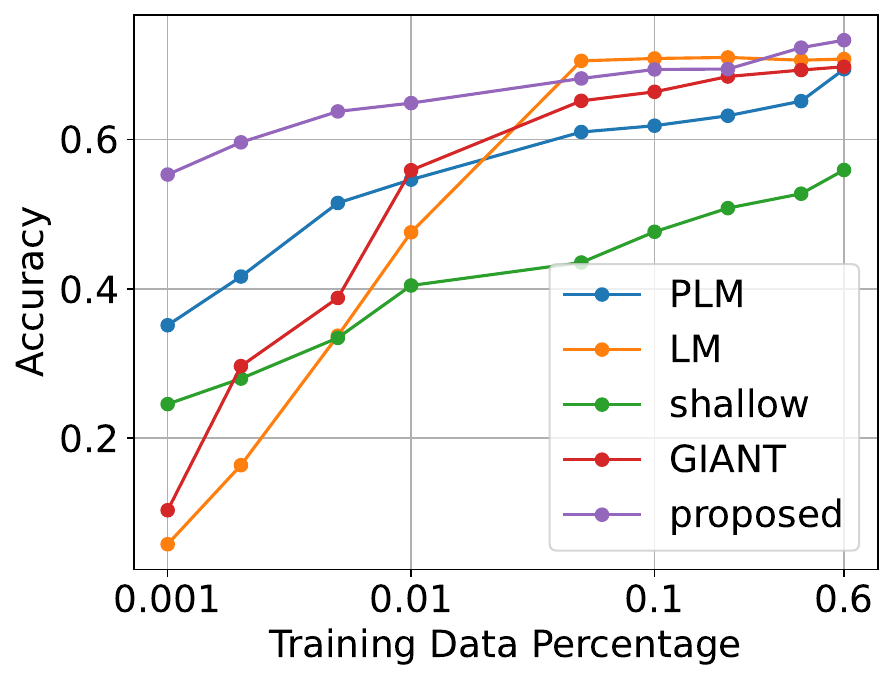}
        \caption{ogbn-Products}
    \end{subfigure}
    \begin{subfigure}{0.24\textwidth}
        \includegraphics[width=\linewidth]{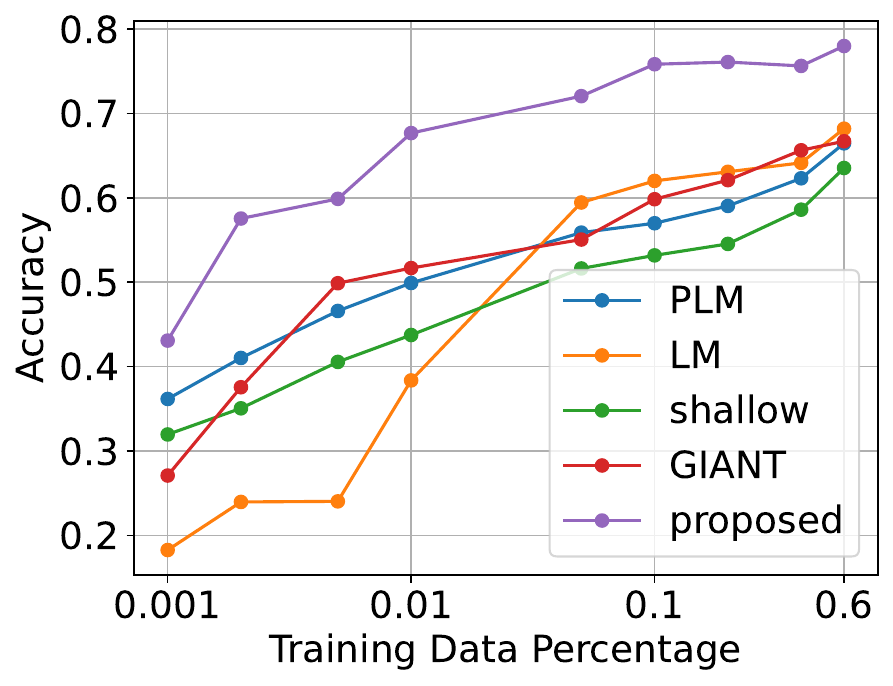}
        \caption{Arxiv-2023}
    \end{subfigure}
    \vspace{-4mm}
    \caption{Test accuracy on different proportions of available training data on four datasets. The x-axis represents the percentage of the training data, ranging from 0.001 to 0.6 of the dataset, plotted in the logarithm scale. The y-axis represents the test accuracy.}
    \label{fig:data-efficiency}
\end{figure*}

\subsection{Main Results of Label-Free Node Classification} \label{sec:main-results}
We first compare our proposed method with other test-time LLM-free TAG learning methods. The main results are shown in Table~\ref{tab:main}. Generally, our proposed framework achieves the best performance on all datasets. More specifically, we have the observations as follows.

\textbf{Best performance on all datasets.} Our proposed method achieves the highest accuracy on all datasets, demonstrating its general efficacy. Compared to the second-best scores, our proposed method yields an over $10.3\%$ performance improvement on Cora, $2.2\%$ improvement on PubMed, $4\%$ improvement on ogbn-products and $8.3\%$ improvement on arxiv-2023, showing our proposed method can consistently beat the all the comparison methods with significant improvements. Especially, Arxiv \cite{he2023explanations} is a dataset that ranges out of the training data of GPT-3.5 Turbo, which stresses the concern that the performance increase may due to popular datasets are seen in LLM's training stage.



\textbf{LM-based methods beat traditional GNN-based methods on the label-free setting.} Among our comparison methods, we found pure LM methods achieve performance better than some of the GNN-based methods like GNN (PLM) and GIANT, and yet comparable performance as SIMTEG. This observation is different from the traditional supervised learning setting, where GNN-based methods typically perform better. This discrepancy is possibly due to the accumulation of inaccuracies in LLM's pseudo-labels within GNN-based methods. The results suggest that when using LLM's pseudo labels for training (with a larger noise level), LM-based methods can give a more robust performance. Our method, utilizing the LLM's information as well as the semantic rationales, also benefits from such capability.

\textbf{Potential as pre-training in standard supervised learning.} We also demonstrate that our proposed framework can be used as an effective model pre-training method when the downstream task is a standard supervised learning setting on the same data. We conducted experiments on Cora dataset to validate the performance increase. The results are shown in Table.~\ref{tab:pretrain}. Results reveal that the proposed distillation can serve as an effective model pre-training in the supervised learning setting. 
It is worth noting that this pre-training uses the same set of TAG data as the supervised learning setting.


\begin{table}
\small
\centering
\caption{Performance comparison between standard supervised learning and using proposed method as model pre-training on Cora dataset.}
\label{tab:pretrain}
\resizebox{0.48\textwidth}{!}{
\begin{tabular}{lccccc}
\toprule
Method & GCN & GAT & SAGE \\
\midrule 
w/o Pre-training & 0.8778±0.0084 & 0.8750±0.0137 & 0.8815±0.0093 \\
w/ Pre-training & 0.8882±0.0020 & 0.8813±0.0015 & 0.8910±0.0028 \\
\bottomrule
\end{tabular}}
\end{table}

\subsection{Efficiency Study} \label{sec:efficiency}
To further analyze the efficiency of the proposed method, we conduct an efficiency study focusing on the performance of our method on different proportions of available training data, and the training and test time.

\textbf{Training Data Efficiency.} We conduct the training data efficiency analysis to explore our method's performance when training data is limitedly available. The results are shown in Fig.~\ref{fig:data-efficiency}. Specifically, we adjust the range of the percentage of available training samples in the whole dataset from $0.1\%$ to $60\%$ and plot the test accuracy curves. Results show that our proposed distillation method performs significantly better than all comparison methods when training data is scarce. This reveals distilling the rationale behind predictions can help the model better learn the rule with limited data, validating the effectiveness of our rationale distillation. Results also reveal our method consistently delivers the best performance across almost all training data percentages, although the second-best method may vary. This demonstrates the effectiveness of our approach.

\begin{table}
\centering
\caption{Computational cost comparison of different methods. \#tokens column denotes the total number of tokens required to pass to and receive from LLMs.}
\label{tab:cost}
\setlength{\tabcolsep}{4pt} 
\renewcommand{\arraystretch}{1.2} 
\begin{tabular}{llcccc}
\toprule
LLM's Role & \#tokens & Method & \makecell{Training\\time} & \makecell{Test\\time} & {Avg. Acc.} \\
\midrule
\multirow{3}{*}{Annotator} & \multirow{3}{*}{1.4M} & LM & 118s & 21s & 0.7380 \\
& & GNN (PLM) & 34s & 20s & 0.6689 \\
& & GIANT & 470s & 21s & 0.7194 \\
\midrule
Teacher & 4.2M & Proposed & 262s & 21s & 0.8228 \\
\bottomrule
\end{tabular}
\end{table}

\textbf{Computational Cost.} As a method designed for LLM-free prediction, we further evaluate the computational cost of our method comparing to other method using LLM-provided labels. Results are shown in Table~\ref{tab:cost}. We evaluate the total number of tokens that require to pass to LLM (\#tokens), the offline training time and test time of the models, and the test accuracy. Results show that although our model needs to pass more tokens to LLMs to get the rationales compared to only getting pseudo-labels, the training and test time is comparable to other methods. Our method achieves significantly better results with the cost of more input and output of LLMs, which is exactly our designation purpose of leveraging more information from LLMs.

\subsection{Ablation Study}\label{sec:ablation}
In the ablation study, we conduct experiments to evaluate 1) each enhanced feature's contribution to the interpreter model, and 2) the improvement of our alignment method compared to the vanilla alignment of soft labels. 

\textbf{Interpreter Model's Feature Enhancements.}
We conduct an ablation study of each enhanced feature on Cora dataset. The results are shown in Table~\ref{tab:ablation-1}. To construct ablated versions, we remove the LLM-generated soft labels as supervision (w/o soft labels), keywords recognition (w/o keywords), key edges recognition (w/o key edges), and key messages (w/o messages), correspondingly. The results validate the effectiveness of each feature enhancement for the interpreter model. 
Specifically, the results show that the proposed method with all components achieves the best average performance (ranked first). This reveals using full components can lead to a more reliable performance. The results also reveal that key edges and LLM-generated messages play a relatively larger role in the interpreter model's performance improvement, which is aligned with the nature of TAGs that structure-related information plays a more important role than pure text information.

\begin{table}

\small
\centering
\caption{Ablation study on feature enhancements, showing interpreter model accuracy. Performances are distinguished: \textbf{best} (bolded), \uline{second}, and \uuline{third}. Includes a \textit{Rank.} column for average accuracy ranking across three GNN backbones.}
\label{tab:ablation-1}
\resizebox{0.48\textwidth}{!}{
\begin{tabular}{lcccc}
\toprule
Method & GCN & GAT & SAGE & Rank. \\
\midrule 
Baseline & 0.8090±0.0174 & 0.8058±0.0320 & 0.8196±0.0294 & 6 \\
w/o soft labels & 0.8310±0.0376 & \uuline{0.8292±0.0451} & $\mathbf{0.8369}$±$\mathbf{0.0527}$ & 3 \\
w/o keywords & \underline{0.8363±0.0291} & \uline{0.8344±0.0288} & 0.8322±0.0312 & 2 \\
w/o key edges & 0.8247±0.0271 & 0.8215±0.0302 & \uline{0.8353±0.0494} & 5 \\
w/o messages & $\mathbf{0.8379}$±$\mathbf{0.0388}$ & 0.8233±0.0410 & 0.8296±0.0538 & 4 \\
Proposed & \uuline{0.8336±0.0288} & $\mathbf{0.8376}$±$\mathbf{0.0209}$ & \uuline{0.8326±0.0210} & \textbf{1} \\
\bottomrule
\end{tabular}}

\end{table}

\begin{table}
\small
\centering
\vspace{-1mm}
\caption{Ablation study of proposed feature discrepancy feature alignment. The method \textit{Interpreter} denotes the performance of the interpreter model, which is the target (upper bound) for the student model to emulate through alignment.}
\label{tab:ablation-2}
\resizebox{0.48\textwidth}{!}{
\begin{tabular}{lccccc}
\toprule
Method & GCN & GAT & SAGE \\
\midrule 
Interpreter & 0.8336±0.0288 & 0.8376±0.0209 & 0.8326±0.0210 \\
\midrule 
Baseline & 0.8090±0.0174 & 0.8058±0.0320 & 0.8196±0.0294 \\
Proposed-T & 0.8127±0.0235 & 0.8104±0.0157 & 0.8187±0.0307 \\
Proposed-N & 0.8146±0.0196 & 0.8081±0.0116 & 0.8204±0.0304 \\
Proposed & $\mathbf{0.8237}$±$\mathbf{0.0187}$ & $\mathbf{0.8237}$±$\mathbf{0.0137}$ & $\mathbf{0.8210}$±$\mathbf{0.0296}$ \\
\bottomrule
\end{tabular}}
\end{table}

\textbf{Ablation Study of Proposed Model Alignment.} 
We further study the effectiveness of two proposed model alignment terms on the Cora dataset. The results are shown in Table~\ref{tab:ablation-2}. In this experiment, the baseline method (denoted by "Baseline") is vanilla model alignment by minimizing the soft predictions, "Proposed-T" and "Proposed-N" means removing the term of semantics and structure alignment, correspondingly. Results show that the proposed method consistently beat all the ablation versions, thus validating the effectiveness of the two proposed alignment terms. Also, the narrower performance gap between the student model and the interpreter model shows our proposed framework can well infer unseen data with a more consistent performance compared to when LLMs are available.

\subsection{Parameter Sensitivity Analysis} \label{sec:sensitivity}
\begin{figure}
    \centering \includegraphics[width=1\linewidth, page=1]{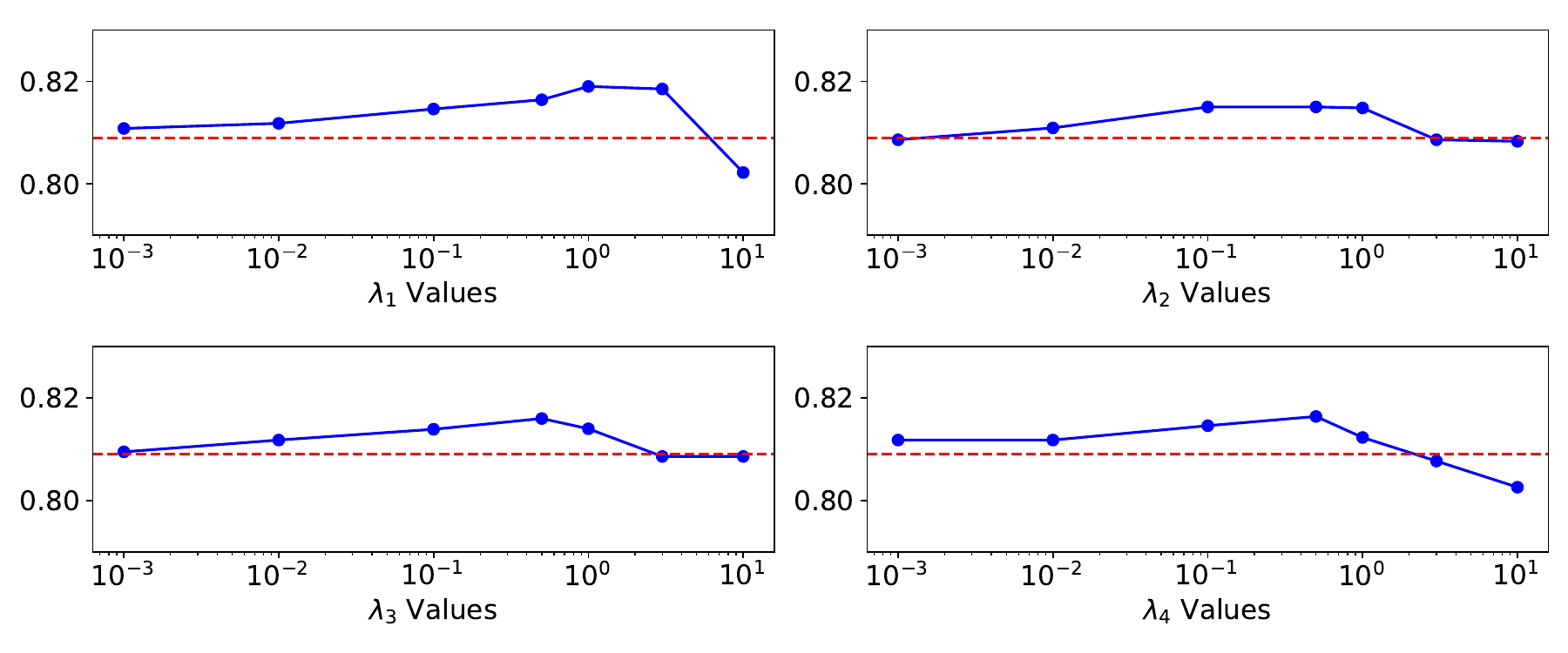}
\vspace{-5mm}
    \caption{Sensitivity analysis of parameters $\lambda_1$, $\lambda_2$, $\lambda_3$ and $\lambda_4$. The red line denotes the baseline performance.}
    \label{fig:sensitivity}
\end{figure}

We explore the sensitivity of the hyper-parameters in our framework, namely $\lambda_1$, $\lambda_2$, $\lambda_3$, $\lambda_4$. For each hyper-parameter, we train the model on seven values ranging from 0.001 to 10, and test the performance of node classification accuracy. Experiments are done on Cora dataset with GCN as the model backbone. The results are shown in Fig.~\ref{fig:sensitivity}. 
Generally, the results demonstrated the robustness of the proposed framework against the varying of its hyper-parameters. To be more specific, the two parameters corresponding to our proposed alignment terms, $\lambda_2$ and $\lambda_3$, show a higher stability of the model performance against the change of their values. That shows the stableness of our proposed model alignment method. The parameter $\lambda_1$, which corresponds to the loss calculated between the interpreter model-predicted soft labels and LLM-generated soft labels, shows a large performance drop when its value reaches 10. This can be explained as the LLM-generated soft labels are not fully accurate.
\section{Conclusion}\label{sec:conclusion}
In this paper, we propose a novel framework to distill LLMs to graph models on TAG learning which allows training with both predictions and rationales. We convert the text rationales to multi-level graph rationales and train an interpreter model to bridge LLM's rationales to the graph model and align the interpreter model based on the nature of TAG data. On four datasets, our proposed distillation method overperforms the baseline by an average of 6.2\%. 

\section*{Acknowledgements}
This work is supported by the NSF Grant No. 1755850, No. 1841520, No. 1942594, No. 2403312, No. 2007716, No. 2007976, No. 1907805.
\bibliographystyle{ACM-Reference-Format}
\balance
\bibliography{rref}

\end{document}